\documentclass[10pt,twocolumn,letterpaper]{article}

\usepackage{cvpr}              

\usepackage{times}
\usepackage{epsfig}
\usepackage{graphicx}
\usepackage{amsmath}
\usepackage{amssymb}

\usepackage{subcaption}
\usepackage[table]{xcolor}
\usepackage{svg}
\usepackage{booktabs}
\usepackage{comment}
\usepackage{multicol}
\usepackage{siunitx}
\usepackage{multirow}
\usepackage[normalem]{ulem}

\definecolor{ForestGreen}{RGB}{34,139,34}
\definecolor{Dandelion}{RGB}{253,188,66}

\newcommand{\method}{InstantAvatar}
\newcommand{\decoder}{\mathcal{F}_\theta}
\newcommand{\renderfunc}{\mathcal{G}_\phi}
\newcommand{\appearnet}{\mathcal{Q}_\rho}
\newcommand{\rendernet}{\mathcal{R}_\eta}

\newcommand{\pedecoder}{\gamma_\mathcal{F}}
\newcommand{\peappearnet}{\gamma_\mathcal{Q}}
\newcommand{\perendernet}{\gamma_\mathcal{R}}

\newcommand{\bg}{\mathbf{g}}

\newcommand{\bl}{\mathbf{l}}
\newcommand{\bn}{\mathbf{n}}

\newcommand{\bv}{\mathbf{v}}

\newcommand{\bx}{\mathbf{x}}

\newcommand{\bz}{\mathbf{z}}

\newcommand{\argmin}{\operatornamewithlimits{arg\,min}}

\newcommand{\mL}{\mathcal{L}}

\newcommand{\mS}{\mathcal{S}}

\usepackage[pagebackref,breaklinks,colorlinks]{hyperref}

\def\paperID{56} 
\def\confName{3DV\xspace}
\def\confYear{2024\xspace}

\title{InstantAvatar: Efficient 3D Head Reconstruction via Surface Rendering}

\author{
Antonio Canela $^{1,2,3}$\hspace{0.2cm}
Pol Caselles $^{1,2,3}$\hspace{0.2cm}
Ibrar Malik $^{1,2,3}$\hspace{0.2cm}
Eduard Ramon $^{1,2}$\thanks{This work was done prior to joining Amazon.}
\\
Jaime García $^{1}$\hspace{0.4cm}
Jordi Sànchez-Riera $^{3}$\hspace{0.4cm}
Gil Triginer $^{1}$\hspace{0.4cm}
Francesc Moreno-Noguer $^{3}$
\and
$^1$Crisalix Labs \hspace{4mm} $^2$Universitat Politècnica de Catalunya \\
$^3$Institut de Robòtica i Informàtica Industrial, CSIC-UPC
}

\begin{document}
\maketitle


\begin{abstract}
Recent advances in full-head reconstruction have been obtained by optimizing a neural field  through differentiable surface or volume rendering to represent a single scene. While these techniques achieve an unprecedented accuracy, they take several minutes, or even hours, due to the expensive optimization process required. In this work, we introduce InstantAvatar, a method that recovers full-head avatars from few images (down to just one) in a few seconds on commodity hardware. In order to speed up the reconstruction process, we propose a system that combines, for the first time, a voxel-grid neural field representation with a surface renderer. Notably, a naive combination of these two techniques leads to unstable optimizations that do not converge to valid solutions. In order to overcome this limitation, we present a novel statistical model that learns a prior distribution over 3D head signed distance functions using a voxel-grid based architecture. The use of this prior model, in combination with other design choices, results into a system that achieves 3D head reconstructions with comparable accuracy as the state-of-the-art with a 100$\times$ speed-up.
\end{abstract}

\vspace{-2mm}


\vspace{-2mm}

\section{Introduction}
\label{sec:intro}

 Obtaining 3D avatars from images or videos is an essential building block for many mixed reality applications. Traditional methods based on 3D morphable models (3DMM) \cite{bfm09, li2017learning} have proved to be a robust solution, but they struggle to capture details outside of the training distribution, and cannot handle topological changes caused by hair, clothes and other accessories. Voxel grids provide more flexibility than 3DMM to represent arbitrary shapes, but they are limited by their memory requirements. In order to avoid these limitations, neural fields have been proposed as a highly expressive 3D representation at a fixed memory footprint.

 In combination with neural rendering, neural fields have achieved impressive results for the tasks of novel view synthesis \cite{mildenhall2020nerf} and 3D reconstruction \cite{yariv2020multiview}, and have been successfully applied to represent full heads with unprecedented accuracy and photorealism \cite{Ramon_2021_ICCV, kellnhofer2021neural, caselles2022sira}. More concretely, methods that represent the geometry implicitly using a Signed Distance Function (SDF) have led to very detailed surface reconstructions, either when optimized through surface  \cite{yariv2020multiview} or volume rendering \cite{yariv2021volume, oechsle2021unisurf, wang2021neus}. These methods have been applied to recover full-head avatars from videos \cite{zheng2022avatar, grassal2022neural} or still photos \cite{Ramon_2021_ICCV,caselles2022sira,Chatziagapi_3dv2021}, achieving more detailed and visually appealing reconstructions than their 3DMM-based alternatives.

\begin{figure}[t]
\vspace{-2mm}
\centering
    \captionsetup{type=figure}
\includegraphics[width=0.99\columnwidth]{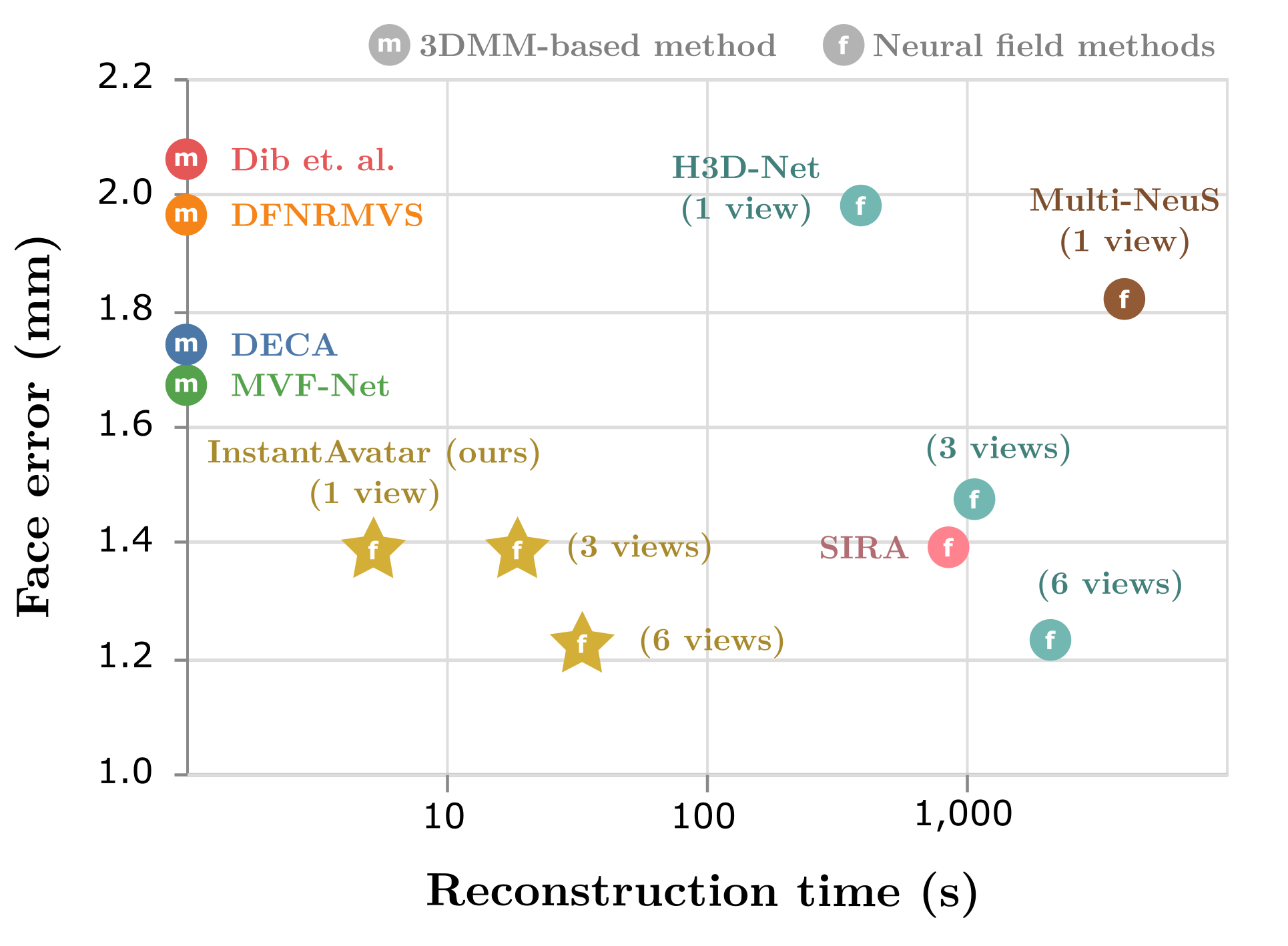}
\vspace{-3mm}
     \caption{\textbf{Reconstruction time comparison}. InstantAvatar is a method that obtains full head 3D avatars from one or few images in a matter of seconds. This figure reports the  time vs reconstruction error for ours and state-of-the-art methods, when considering only the face region  where all methods are applicable (full head error metrics are reported at the experiments section). InstantAvatar speed is only surpassed by 3DMM methods, which, however are significantly less accurate. Compared against other neural field approaches, InstantAvatar obtains a 100$\times$ speed up at similar reconstruction error values. 
 }
     \label{fig:teaser}
\label{fig:teaser}
\vspace{-3mm}
\end{figure}

However, techniques based on differentiable rendering take several minutes, or even hours, to reconstruct a single scene due to the costly optimization process involved. To alleviate this, some works have leveraged hybrid architectures that encode neural fields as a composition of a feature grid and a shallow multi-layer perceptron (MLP) \cite{Takikawa_2021_CVPR, muller2022instant, sun2022direct} and other techniques in order to accelerate these processes \cite{neus2, li2023nerfacc, yu2022plenoxels, yu2021plenoctrees}. These hybrid architectures have led to much faster optimizations of neural fields through volume rendering. Still, the convergence time for these approaches remains in the tens of minutes \cite{li2022vox}, which is prohibitively high for many applications.

A promising direction towards further speeding up the optimization of neural fields through differentiable rendering is optimizing a hybrid neural architecture through surface rendering instead of volumetric rendering. The potential speed gains arise because volume rendering is slowed down by the computation of both a forward and a backward pass for a large number of points along rendered rays. Unfortunately, combining hybrid neural architectures and surface rendering is far from trivial. We observe that the lack of inductive biases for smoothness in a feature grid, together with the highly sparse sampling from the surface rendering leads to an unstable optimization process that does not converge to valid solutions. This problem is accentuated when the number of input images is reduced.

In this paper, we propose a method for successfully optimizing hybrid neural architectures through surface rendering, and apply it to the challenging task of full head 3D reconstruction from few input images, down to a single image. In this case, our main objective is to achieve a significant speed-up while obtaining competitive results on geometry reconstruction. Thus, we keep the rendering setup as presented in previous papers \cite{Ramon_2021_ICCV, yariv2020multiview}. We begin by training a multi-resolution grid-based neural field to represent a statistical prior of head SDFs, and then use it during the 3D reconstruction process. We avoid dual surface/volume rendering techniques \cite{yariv2021volume, oechsle2021unisurf, wang2021neus}, and use direct surface rendering \cite{yariv2020multiview} instead. We show that, while a naive optimization would fail to converge, using a statistical prior allows us to reliably arrive to a coarse solution, which we later refine. To further accelerate and stabilise the reconstruction convergence, we include normal cues as supervision, leveraging the predictions of a monocular normal estimator\cite{Yu2022MonoSDF}. In summary, our main contributions are:

\begin{itemize}
  \setlength\itemsep{0.1em}
    \item We introduce, for the first time, a framework that combines a grid-based architecture with a surface rendering method that yields to fast and accurate 3D reconstructions from one or few input images.
    \item We  leverage on a statistical prior, obtained with thousands of 3D head models, to guide network convergence and achieve a reconstruction accuracy on a par with state of the art methods, but  with $\sim$100$\times$ speed-up.
    \item We provide an optimal training scheme for grid-based structures combined with surface rendering methods exhibited through a variety of datasets evaluated.
\end{itemize}

\vspace{-2mm}

\section{Related work}
\label{sec:related-work}

\begin{figure*}[!t]
    \vspace{-7mm}
    \centering
    \includegraphics[width=\textwidth]{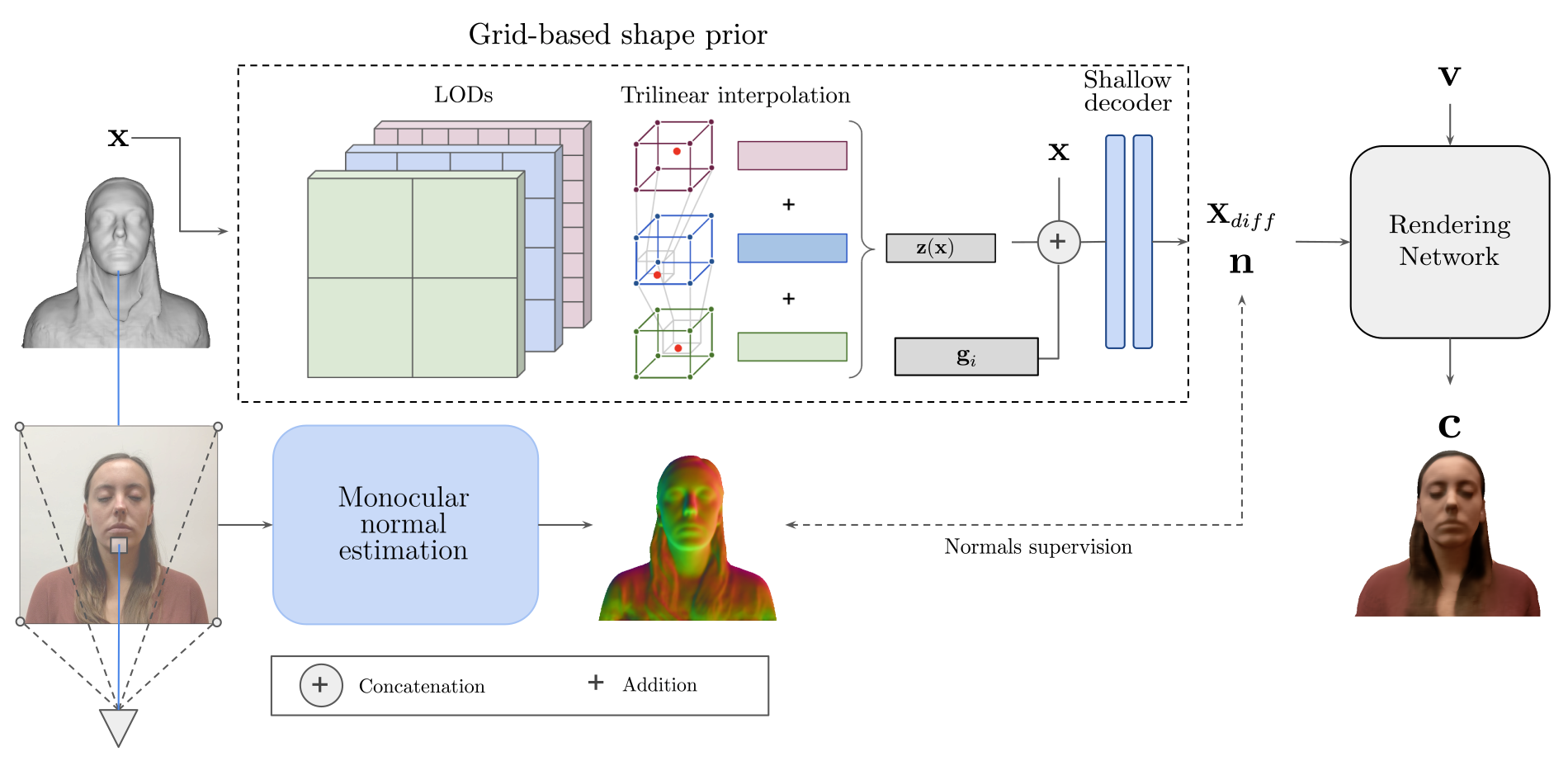}
    \caption{\textbf{Overview of our method.} For each query point $\bx$ we obtain the feature $\bz(\bx)$ from the multi-resolution feature grid at different levels of detail. Afterwards, we concatenate the positional encoding applied to $\bx$, the global feature $\bg_{i}$, and the grid feature $\bz(\bx)$ to query the SDF parameterized by a shallow MLP. We supervise the gradient of the SDF with the predicted normal at the pixel location where the ray intersects. Finally, we use a rendering network to predict the radiance emitted from the surface point $\bx_{diff}$, with normal $\bn$, in a viewing direction $\bv$.}
    \label{fig:method}
    \vspace{-4mm}
\end{figure*}

Reconstructing objects in 3D from very few in-the-wild images, or even a single one, is a highly unconstrained problem that requires prior knowledge in order to be solved. This typically comes in   the from  statistical models \cite{bfm09, FLAME:SiggraphAsia2017, SMPL:2015, wang2022faceverse, Brunton2014_Wavelets, smith2020morphable}.

3D Morphable Models (3DMM) are the most common representation to encode statistical properties of an object category, and have been extensively used for multi-view 3D face reconstruction~\cite{bai2020deep, dou2018multi, ramon2019multi, wu2019mvf} and single view 3D reconstruction ~\cite{richardson20163d, richardson2017learning, tewari2017mofa, tran2018extreme, tuan2017regressing}. However,  while they are fast and robust, they lack flexibility. In order to capture more details, they are combined with differentiable post-processing techniques like normal maps, which capture high-frequency information ~\cite{Lin_2020_CVPR, richardson2017learning, tran2018extreme}. 

Recently, neural fields have been proposed as a more flexible representation for 3D shape modelling \cite{park2019deepsdf,littwin2019deep,mescheder2019occupancy}. In combination with differentiable rendering, neural fields have been successfully applied for the task of 3D object reconstruction from several images \cite{niemeyer2020differentiable, yariv2020multiview}. In order to reduce the amount of required input information, \cite{Ramon_2021_ICCV,caselles2022sira,Yu2022MonoSDF} introduce priors that guide the optimization process towards plausible solutions. While reducing the number of input images also reduces the optimization time, these methods still require several minutes or even hours to reconstruct a single scene. Apart from presenting good results in terms of geometry reconstruction accuracy, some methods also cover face expressions as in \cite{cao2022authentic}, however this method is out of our scope as it uses RGB-D data and its runtime is in a higher order of magnitude.

Several efforts have been devoted to speed up the optimization process of neural fields. \cite{sitzmann2019metasdf, tancik2020meta} propose using meta-learning to reduce the number of steps to optimize a single model. \cite{Lindell20arxiv_AutoInt} learns the integral of volumetric rendering. Other methods propose more efficient architectures that can be queried faster. \cite{lindell2022bacon} proposes an architecture that can use an arbitrary number of sequential layers to query the value of a neural field at expenses of constraining the frequency domain. This enables to use coarse-to-fine optimization schemes, avoiding unnecessary computations at the coarser levels. Similarly, \cite{mueller2022instant,sun2022direct, Yu2022MonoSDF} combine multi-resolution feature grids with shallow MLPs in order to reduce the inference time too, and to adapt resources to the optimization stage. Methods based on multi-resolution feature grids have been successfully used in combination with differentiable volume rendering to obtain 3D reconstructions \cite{Yu2022MonoSDF,li2022vox}. However, volumetric rendering requires sampling several points per ray, making it computationally intense.

 While surface rendering is more efficient than volumetric rendering, since it only requires a single point per ray, it has not yet been applied to reconstruct 3D scenes in combination with multi-resolution feature grids. We believe this is due to the optimization problems derived from using a neural architecture without inductive biases for continuity plus the highly sparse sampling from surface rendering. 
In this paper we build and use an statistical model to enable, for the first time, optimizing a hybrid  neural architecture  with surface rendering.


\vspace{-2mm}

\section{Method}
\label{sec:method}

Our goal is to reconstruct the 3D head of a subject from a small set of images, with associated foreground masks and camera parameters. We parametrize the surface to reconstruct $\mS$ as the zero level iso-surface of a continuous signed distance function (SDF).


To guide the reconstruction process, we build a grid-based shape statistical model that represents a prior distribution over 3D head SDFs. We extend the auto-decoder architecture proposed in \cite{park2019deepsdf} with a multi-resolution 3D feature grid that allows reducing the size of the MLP decoder and speeding up SDF queries. To encourage feature grids of increasing resolution to encode progressively finer geometric details we propose a curated scheduling for the prior training.


During the 3D reconstruction process, our architecture builds on top of H3D-Net \cite{Ramon_2021_ICCV}, using differentiable surface rendering and minimising photoconsistency and silhouette errors. Crucially, at the beginning of the optimization, the SDF is constrained to be within the learned shape prior space. This allows reaching a coarse solution first, avoiding irrecoverable incorrect local minima, after which we can lift the constraint to obtain finer details.

To better guide the optimization, we supervise the gradient of the SDF with the predictions of a monocular normal map estimation model, similar as in \cite{Yu2022MonoSDF}. Inspired by ray sampling in volumetric rendering approaches \cite{mildenhall2020nerf}, we use an algorithm better suited for parallel computing to find the zero level set iso-surface for a given SDF that is up to 30\%  faster than the sphere tracing algorithm provided in \cite{yariv2020multiview}.

We next detail each of the modules of our method. An overview of the method can be found in Figure \ref{fig:method}.

\subsection{Grid-based shape prior}
\label{sec:method_prior}
\noindent
\textbf{Architecture.} Parametrizing a space of signed distance functions with a monolithic MLP decoder is computationally expensive, since any evaluation involves the entire network. To trade compute time for memory, we extend the H3D-Net \cite{Ramon_2021_ICCV} framework with a multi-level dense feature grid, which allows reducing the size of the MLP decoder. To obtain the signed distance $d$ for a query point $\bx \in R^3$ we compute:

\vspace{-5mm}

\begin{equation}
    d = \decoder([\pedecoder(\bx), \bz, \bg])\;,
\label{eq:sdf}
\end{equation}
where $\decoder$ is a shallow MLP decoder, $[\cdot, \cdot]$ denotes concatenation, $\gamma(\cdot)$ denotes sinusoidal positional encoding \cite{tancik2020fourfeat}, and $\bg$ is a latent vector encoding specific shapes in the shape space.

The vector $\bz$ is a feature interpolated from a multi-resolution feature grid. For each level $l \in [4,5,6]$ we define a dense voxel grid $V_l$ with $2^{l*3}$ voxels in a bounding volume of $[-1,1]^3$. We define an associated feature grid $Z_l$ with $(2^{l}+1)^{3}$ features located at the corners of each voxel in $V_l$. Any query point in the bounding volume can be assigned to a voxel for every resolution level, and a feature can be computed by trilinearly interpolating the corner features. To aggregate the features from different levels we sum them:

\vspace{-5mm}

\begin{equation}
    \bz(\bx) = \sum_{l \in L}^{}{{\rm interp}(\bx, Z_l)}\;.
\end{equation}

\vspace{-1mm}

Our shape prior has been trained on a collection of coarse head scans. It consists of non-watertight meshes, indexed by $i=1,\ldots,M$. To learn an SDF from a mesh in this set we will minimize the signed distance on the surface points, and ensure the correctness of the function with an eikonal loss term \cite{igr}. The space shape can be learned from the following objective function \cite{Ramon_2021_ICCV}:
\begin{equation} \label{eq:deep_sdf_objective}
\argmin_{\{\bz, \bg_i\}, \theta} \sum_{i=1}^M \mL_{\rm Surf}^{(i)} + \lambda_0\mL_{\rm Emb}^{(i)} + \lambda_1\mL_{\rm Eik}^{(i)}\;,
\end{equation}
where $\lambda_0$ and $\lambda_1$ are hyperparameters and $\mL_{\rm Surf}$, $\mL_{\rm Eik}$ and $\mL_{\rm Emb}$ account respectively for the SDF error at surface points, the eikonal term as in \cite{gropp2020implicit, yariv2020multiview, Ramon_2021_ICCV, caselles2022sira} and a regularisation applied to latents $\bz(\bx)$ and $\bg_i$. We enforce a zero-mean multivariate Gaussian distribution with spherical covariance $\sigma^2$ over the spaces of shapes: $\mathcal{L}_{\rm Emb}^{(i)} = \frac{1}{\sigma^{2}}\bigl(\lVert \bz(\bx) \rVert_2 + \lVert \bg_i \rVert_2\bigr)$. More details about these losses can be found at the supplementary material.

\vspace{1mm}\noindent\textbf{Training schedule.} We aim to train the shape prior so that feature grids of increasing resolution encode progressively finer geometric details.  
To this end, we propose the following scheduling. We initialise with zeros the feature grids and the decoder using geometric initialization \cite{Atzmon_2020_CVPR}. During a first stage, taking $N$ epochs, we train the coarse grid $Z_4$ along with the decoder $\decoder$ and the global latent $\bg$, leaving the finer feature grids frozen. Once a coarse fit of the scenes has been obtained, we freeze all the parameters except for the next feature grid ($Z_5)$. After $2*N$ epochs, we repeat the same procedure, training only the last level ($Z_6$) during $4*N$ epochs. The progressively longer stages are due to the higher resolution grid, and therefore larger number of parameters, introduced in each stage.

In order to make the training process more stable and promote low frequency shapes to be encoded by coarse levels, we use a progressive masking of the positional encoding $\gamma(\cdot)$ \cite{lin2021barf}. At the beginning of the optimization process all frequencies are masked and it is being unmasked progressively between epochs 0 and $N/2$.

\subsection{Surface reconstruction}

To obtain the 3D reconstruction of a scene given posed images and foreground masks, we follow a surface rendering approach to optimize our pre-trained statistical model, similar as in \cite{Ramon_2021_ICCV}. For every pixel, we march a ray and find the ray-surface intersection point, $\bx$, which we make differentiable with respect to the network parameters through implicit differentiation \cite{yariv2020multiview}, which we denote by $\bx_{diff}$. We then render the object at this point, with normal $\bn$, seen from the ray viewing direction $\bv$, using a differentiable rendering function $\renderfunc(\bx,\bn,\bv)$, with learnable parameters $\phi$. In addition to the usual losses penalising photoconsistency error, and silhouette error, we add a loss term supervising the surface normal vectors with the predictions of a monocular normal map regressor. We use a cosine similarity loss
\begin{equation}
    \mL_{\rm norm} = 1 - \nabla_\bx d \cdot \hat{\bn}\;,
\end{equation}
where $d$ is the SDF value as defined in eq. \ref{eq:sdf}, so its gradient at the surface corresponds to the normal vector, and $\hat{\bn}$ is the prediction of our monocular normal map estimation model.

The scheduling used to optimize \method{} for a specific scene closely follows the one proposed in \cite{Ramon_2021_ICCV}. First, we freeze the weight of the decoder and we train the global feature vector $\bg_i$ and all the feature grids $Z_l$ during 30 epochs. Afterwards, we unfreeze the decoder $\decoder$ until epoch 100.

It is worth mentioning that, taking into account the feature grid architecture being used, we are then not able to make full use of the eikonal equation due to the inter\-voxel non-continuity of the SDF derivatives with respect to $\bx$. In addition, the eikonal loss has minima which are not SDFs as explained in \cite{pumarolavisco}, for this reason, the eikonal regularization is disabled during the surface reconstruction process.

\vspace{1mm}\noindent\textbf{Monocular normals estimation.}  Our architecture consists of a U-Net++ \cite{zhou2018unetplusplus} with an EfficientNet-B1 encoder \cite{pmlr-v97-tan19a} pre-trained on ImageNet classification. We train the model to predict a normal map from a single input image, expressing normals in the camera frame of reference. The training data is generated from a collection of raw head scans paired with posed images, which we use to render ground truth normal maps. We generate 50k pairs of images and normal maps, covering $\pm 90$º from the frontal view. See more details in the supplementary material.

\vspace{1mm}\noindent\textbf{Ray-surface intersection finding algorithm.} We take inspiration from volumetric rendering approaches \cite{mildenhall2020nerf} to accelerate the ray tracing algorithm introduced in \cite{yariv2020multiview}. Instead of using an iterative procedure to find the roots of an SDF, we propose a parallelizable approach more suited for high-end GPUs. First, we sample $N_p$ points along each ray from the pixel towards the bounding volume in a viewing  direction $\bv$. Secondly, we select the first interval between points where the field has a sign change and we sample $N_f$ points in this segment. We return as the intersection coordinate, the first point where the field has a zero crossing. See more details in the supplementary material. 

\subsection{Implementation details}
We now describe the implementation details of our architecture and how we train it.
Before passing the input coordinates $\bx$ to the decoder network, we apply a positional encoding $\pedecoder(\bx)$ with 6 log-linear spaced frequencies. The encoded 3D coordinates are concatenated with the $\bg_i$ global latent vector of size 256, and the feature interpolated from a multi-resolution feature grid $\bz$ of length 8 and set as the input to the decoder. The decoder $\decoder$ is an MLP of 3 layers with 512 neurons in each layer. We use Softplus as activation function in every layer except for the last one, where no activation is used. As in \cite{Ramon_2021_ICCV}, the rendering function $\renderfunc(\bx,\bn,\bv)$ is implemented as a composition of two sub-networks $\appearnet$ and $\rendernet$. $\appearnet(\peappearnet(\bx))$ is an MLP of 8 layers with 512 neurons in each layer and a single skip connection from the input of the network to the output of the 4th layer. We use Softplus as activation function at every layer except for the last, where no activation is used. The $\appearnet$ output is a 256-dimensional vector $\bl$ that is concatenated to $\perendernet(\bx)$ and $\bn$ and provided to $\rendernet$ as input. As in \cite{yariv2020multiview}, $\rendernet(\perendernet(\bx),\bn,\bl,\bv)$ is an MLP composed by 4 layers, each 512 neurons wide, no skip connections, ReLU activations for every layer except for the output layer which has tanh and outputs the 3-dimensional RGB color. We also apply the positional encodings $\peappearnet$ and $\perendernet$ to $\bx$ with 6 and 4 log-linear spaced frequencies respectively. The prior is trained for 700 epochs in total (MLP, global latent and $Z_4$ during the first 100 epochs, then $Z_5$ during 200 epochs, and lastly $Z_6$ for 400 epochs), using Adam \cite{kingma2014adam} with standard parameters, learning rate of $10^{-4}$ and learning rate step decay of 0.5 every 50 epochs. The value parameterizing the length of the different stages of the prior training is $N=100$, so the entire training takes $7N = 700$ epochs. The shape prior training takes approximately 24 hours for a dataset of 10k scenes. 



\vspace{-2mm}

\section{Experiments}
\label{sec:experiments}
\definecolor{rowblue}{RGB}{220,230,240}


In this section we evaluate \method{}  in   single and multi-view settings, and report both qualitative and quantitative results. We compare against the 3DMM-based methods: DECA~\cite{feng2021learning}, Dib et al.~\cite{dib2021towards}, DFNRMVS~\cite{bai2020deep}, MVF-Net~\cite{wu2019mvf} and against neural fields methods: Multi-NeuS~\cite{burkov2022multi}, H3D-Net~\cite{Ramon_2021_ICCV} and SIRA~\cite{caselles2022sira}. 
We conduct these experiments on the following public datasets: H3DS~\cite{Ramon_2021_ICCV}, 3DFAW~\cite{pillai20192nd} and CelebA-HQ~\cite{karras2017progressive}.


\subsection{Datasets}

\noindent{\bf Training.} We build the probabilistic shape prior, and train the monocular normals prediction model, with the same dataset used in \cite{Ramon_2021_ICCV} and \cite{caselles2022sira}. It is made of 10k scenes composed of 3D head scans paired with multi-view posed images. The dataset is balanced in gender and diverse in age and ethnicity. 

\vspace{1mm}
\noindent{\bf 3DFAW.} This dataset contains videos of human heads paired with 3D reconstructions of the facial area. It also contains high resolution photos taken with a professional camera. We select the same 10 cases from the low-resolution set as in \cite{Ramon_2021_ICCV}, and 17 subjects from the high-resolution set provided by \cite{dib2021towards}, for comparison purposes.

\vspace{1mm}
\noindent{\bf H3DS.}  It consists of 23 human head scenes with multi-view posed images, masks, and full-head 3D textured scans. The dataset consists of 13 men and 10 women.

\vspace{1mm}
\noindent{\bf CelebA-HQ.} High-quality version of CelebA that consists of 30k in-the-wild frontal images at 1024×1024 resolution. At Figure \ref{fig:qualitative_celeb} we have selected a subset of 8 scenes that are diverse in gender and ethnicity.

\subsection{Experimental setup} 
For every evaluation dataset, we align the ground truth and the reconstructions using manually annotated facial landmarks. We refine this alignment by performing ICP \cite{besl1992method} of the ground truth face region to the reconstruction. The face region is defined by all the vertices falling under a sphere of radius 95mm centered at the nose. 

After aligning the meshes, we over-sample points from the reconstructions, rejecting points that are too close to each other. We use the unidirectional Chamfer distance from the defined regions on the ground truth to the predicted reconstructions.

\subsection{Ablation study}

\begin{figure*}[t]
    \vspace{-3mm}
    \centering
    \includegraphics[width=0.98\textwidth]{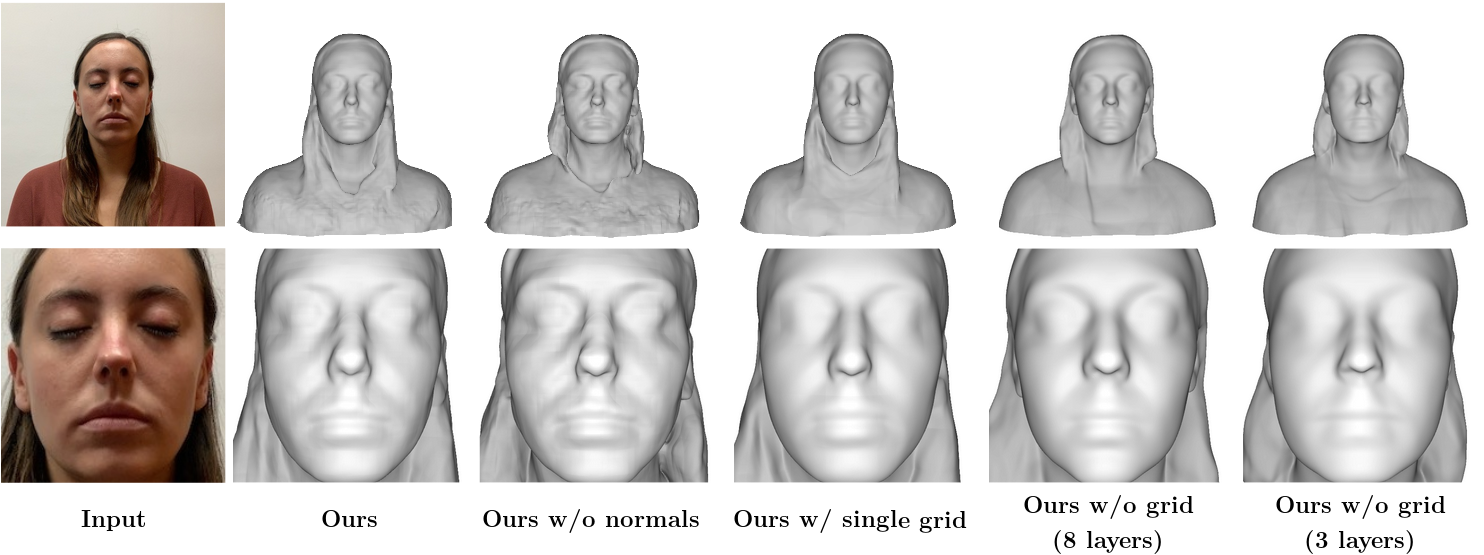}
    \caption{\textbf{Ablation: Qualitative comparison.} We conduct an ablation study to qualitatively compare variations of our model using a H3Ds dataset scene in the multi-view setting (6 views). The bottom row zooms into the face region to better appreciate the differences among configurations. Both our final approach and the one without normals supervision outperform the rest of alternatives. However, when normals supervision is not considered the resulting shape tends to be excessively sharp (e.g. the outermost part of the eyebrows) or erroneous (hair). The single grid and the  8-layer MLP (without grid) results are comparable, although they  are both unable  to capture the high-frequency details obtained with our final model.
    }
    \label{fig:qualitative_ablation}
    \vspace{-3mm}
\end{figure*}

\begin{figure}[]
    \vspace{-4mm}
    \centering
    \includegraphics[width=\columnwidth]{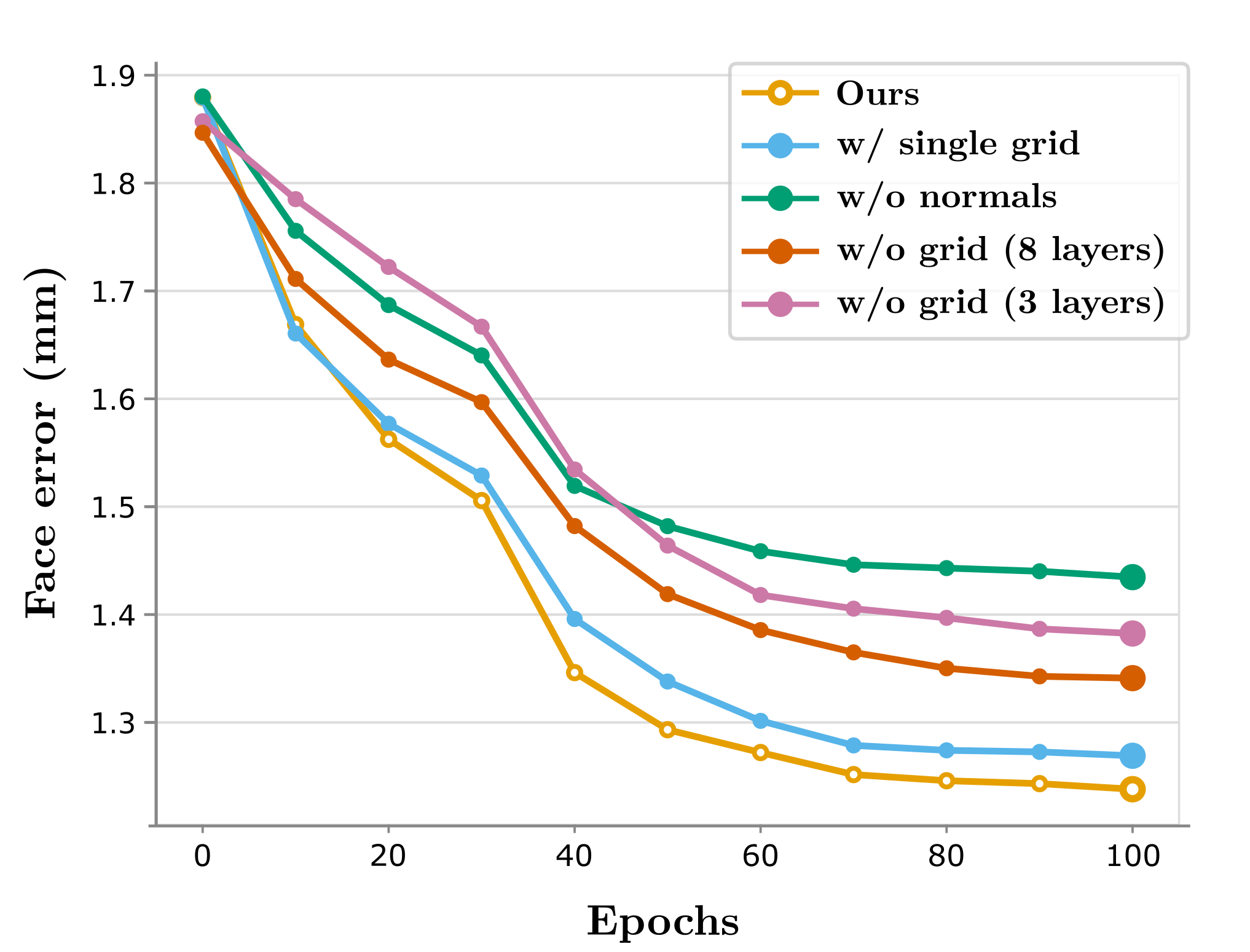}
    \caption{\textbf{Ablation: Convergence speed.} Convergence speed across epochs averaged over all cases of the H3Ds dataset in the multi-view setting (6 views). Grid-base methods show a faster convergence in contrast to MLPs approaches. The supervision of normals helps convergence in the fine-tuning process at the reconstruction stage.}
    \label{fig:conv}
    \vspace{-7mm}
\end{figure}

We conduct an ablation study, evaluating variations of our method in the multi-view reconstruction setting (3 and 6 views) on all cases of the H3DS dataset. We show the qualitative results in Figure \ref{fig:qualitative_ablation} and the quantitative results in Figure \ref{fig:conv} and Table \ref{table:ablation1}. 

A key ingredient to optimize a grid-based SDF with differentiable rendering is to build a prior distribution over 3D head shapes. We observe that when using our architecture without shape prior (see Table \ref{table:ablation1}), the per-scene reconstruction does not converge to plausible human head.

\begin{table}[t]
    \centering
    \setlength{\tabcolsep}{4pt} 
    \vspace{1mm}
    \centering
    \rowcolors{3}{rowblue}{white}
    \resizebox{0.75\columnwidth}{!}{
    \begin{tabular}[t]{lccc}
    \toprule
&  \multicolumn{1}{c}{3 views} &  \multicolumn{1}{c}{6 views} \\
\cmidrule(lr{0.5em}){2-2} \cmidrule(lr{0.5em}){3-3}
                                    & face $\downarrow$ & face $\downarrow$ \\
    \midrule

    Ours w/o shape prior            & 6.98              & 20.72             \\
    Ours w/o normals                & 1.59              & 1.44              \\
    Ours w/o grid (3 layers)        & 1.55              & 1.38              \\
    Ours w/o grid (8 layers)        & 1.46              & 1.34              \\
    Ours w/ single grid             & \textbf{1.39}     & 1.27              \\
    \textbf{Ours}                   & \textbf{1.39}     & \textbf{1.22}     \\
    
    \bottomrule
    \end{tabular}
    }
    \caption{{\bfseries Ablation: Quantitative comparison.}  Each layer of the MLPs are composed of 512 neurons.  Error in millimeters. Average over all cases of the H3Ds dataset in the multi-view setting (3 and 6 views).}
    \label{table:ablation1}
    \vspace{-2mm}
\end{table}

\begin{table}[t]
    \centering
    \setlength{\tabcolsep}{4pt} 
    \vspace{1mm}
    \centering
    \rowcolors{7}{rowblue}{white}
    \resizebox{1.0\columnwidth}{!}{
    \begin{tabular}[t]{lccc}
    \toprule
&  \multicolumn{1}{c}{1 view} &  \multicolumn{1}{c}{3 views} &  \multicolumn{1}{c}{6 views} \\
\cmidrule(lr{0.5em}){2-2} \cmidrule(lr{0.5em}){3-3} \cmidrule(lr{0.5em}){4-4}
                                                      & face $\downarrow$ & face $\downarrow$ & face $\downarrow$ \\
    \midrule

    (A) Ours w/o progressive PE masking               & 1.77              & 1.48              & 1.33              \\
    (B) Ours w/o progressive LODs schedule            & 1.70              & 1.54              & 1.35              \\
    (C) \textbf{Ours}                                 & \textbf{1.50}     & \textbf{1.39}     & \textbf{1.22}     \\
    
    \bottomrule
    \end{tabular}
    }
    \caption{{\bfseries Shape prior training schedules ablation.} Reconstruction metrics (unidirectional Chamfer distance from ground-truth  to predicted reconstructions) using different  prior schedules. We compare our final shape prior schedule (C), which progressively optimizes the levels of detail from coarse to fine, with another schedule that optimizes every level at the same time (B). We also provide an ablation where we have removed the progressive positional encoding masking (A), therefore is fully operative from the start.}
    \label{table:ablation_shape_prior}
    \vspace{-7mm}
\end{table}

When training \method{} without a grid, we observe that the decrease in model capacity leads to a decrease in the capacity to represent high-frequencies, which particularly affects the face region. As seen in Figure \ref{fig:qualitative_ablation}, this has a big impact in the ability to preserve the identity of the subject. To validate it, we have trained a 3-layer and 8-layer MLPs (each layer composed of 512 neurons) without grids and with predicted normals supervision. We have used these two MLP sizes in order to explore the model capacity-speed tradeoff and compare both to our final approach. As shown in Figure \ref{fig:conv}, the multi-grid based method (ours) is the one that converges faster. 

As shown in Table \ref{table:ablation1}, we obtain the smaller error in the face metric when we introduce normal cues as supervision for the gradient of the SDF during the optimization process. We have done an extensive hyperparameter space exploration, and confirmed that the enhanced robustness provided by the normals ensures consistent accuracy. Using a single coarse grid provides robust reconstructions but lacks detail, in contrast, the multi-grid approach is able to represent better the high-frequency details. However, this improvement in the high-frequency domain may not be fully reflected at the shown metrics (Table \ref{table:ablation1}) and would be part of potential future work, as it is an ongoing research subject \cite{REALY}.

In Table \ref{table:ablation_shape_prior}, we compare the performance of different shape prior training schedules, where our final approach with both progressive levels of detail schedule and progressive positional encoding masking outperforms the ablated setups.

\begin{figure*}[t]
    \vspace{-4mm}
    \centering
    \includegraphics[width=0.98\textwidth]{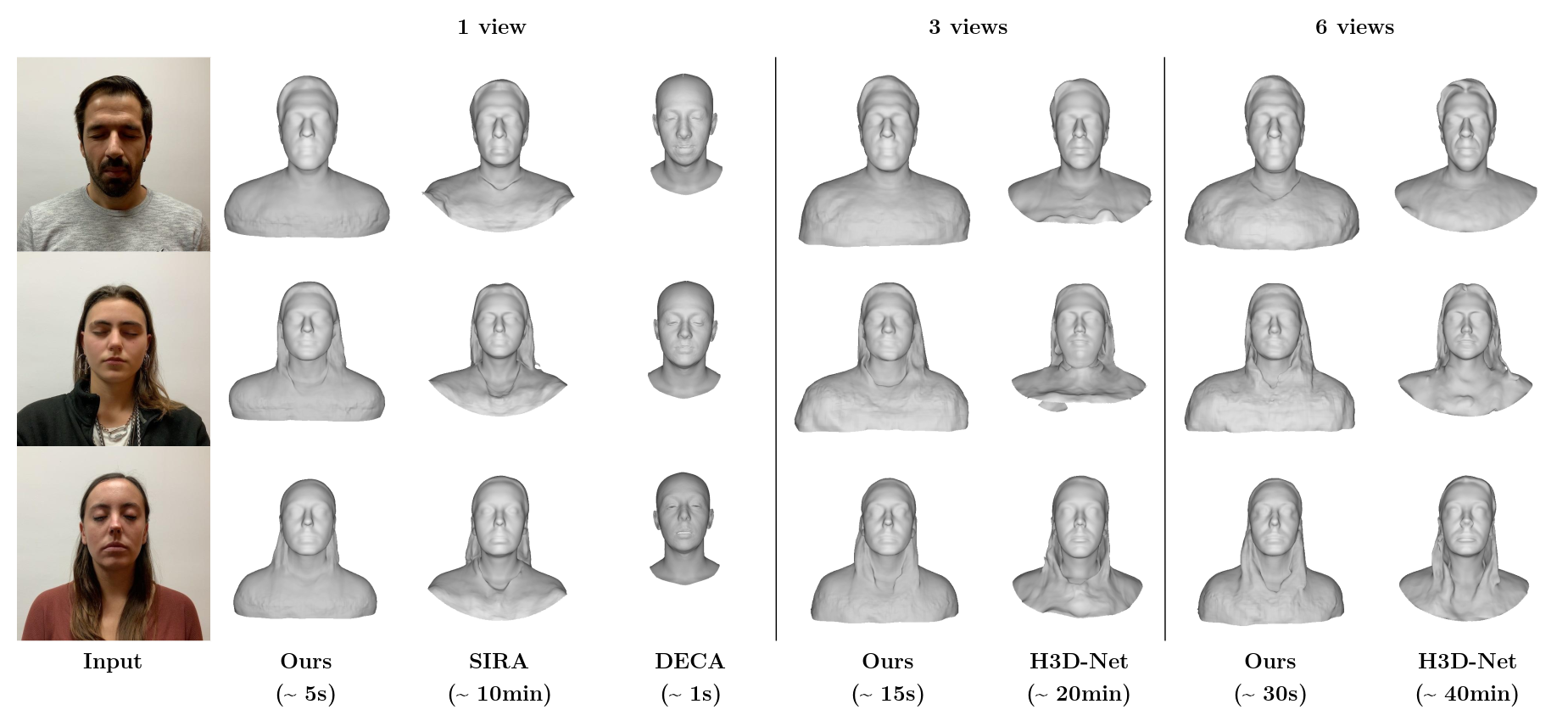}
    \caption{\textbf{Qualitative results.} We compare qualitatively \method{} with other state of the art methods on H3DS dataset for 1 view, 3 views and 6 views. \textbf{1 view:} SIRA is able to better capture the identity of the subject, however, it takes 10 min of training. DECA on the other hand, can only predict the face region. \textbf{3 views: } H3D-Net achieves good bias but at a high variance where we can clearly see artifacts on the chin and the hair. \textbf{6 views: } H3D-Net is able to recover the hair and face regions with similar quality as \method{}.}
    \label{fig:qualitative_experiments_front}
    \vspace{-4mm}
\end{figure*}

\vspace{-1mm}

\subsection{Quantitative results}  

\vspace{1mm}
\noindent{\bf Single-view reconstruction.} We perform reconstructions from front facing images, and report the surface error in Table \ref{table:quantitative1}. We see that our method achieves competitive error metrics, while being more than two orders of magnitude faster than  state-of-the-art methods based on differentiable rendering. More specifically, \method{} is the only method capable of performing full-head reconstruction, including hair and garments, in seconds.

\begin{table}[!htb]
\setlength{\tabcolsep}{4pt} 
\centering
\rowcolors{9}{rowblue}{white}
\resizebox{1.0\columnwidth}{!}{
\sisetup{detect-weight=true}
\begin{tabular}{lcccccccccccc}
\toprule
&  \multicolumn{5}{c}{1 view} \\
\cmidrule(lr{0.5em}){2-6}
& 3DFAW & 3DFAW HR & \multicolumn{2}{c}{H3DS} &   \\
\cmidrule(lr{0.5em}){2-2} \cmidrule(lr{0.5em}){3-3} \cmidrule(lr{0.5em}){4-5}
& face $\downarrow$ & face $\downarrow$ & face $\downarrow$ & head $\downarrow$ &  time $\downarrow$  \\
\cmidrule{1-6}
    Dib et al.                  & 1.99  & 2.16  & 2.06  & -     & \textbf{$\sim$ 1s}     \\
    DECA                        & 1.57  & \textbf{1.43}  & 1.74  & 22.22  & \textbf{$\sim$ 1s}     \\
    \cmidrule{1-6}
    \cmidrule{1-6}
    Multi-NeuS                  & -              & -      & 1.81           & 13.92           &  $\sim$ 1h    \\
    H3D-Net                     & 1.63           & 1.47   & 1.97           & 14.36           & $\sim$ 10min     \\
    SIRA                        & \textbf{1.36}  & 1.51   & \textbf{1.39}  & 14.43           & $\sim$ 10min     \\
    \textbf{Ours}               & 1.58           & 1.69   & 1.50           & \textbf{12.43}  & \textbf{$\sim$ 5s}     \\
\bottomrule
\end{tabular}
}
\caption{{\bfseries 3D reconstruction method comparison (single-view).} Error in millimeters. Time in seconds, approximate upper bound. Methods above the division line are 3DMM-based and methods below are model-free.}
\label{table:quantitative1}
\vspace{-1mm}
\end{table}

\vspace{1mm}
\noindent{\bf Multi-view reconstruction.} In this comparison, the reconstructions are done with 3 and 6 views taken from a similar distance and height. For 3 views, we use a frontal image directly looking at the head and two views placed at $\pm45$ yaw angle. For 6 views, we additionally use two lateral views at $\pm90$ degrees, and a frontal image but with a lower pitch angle.

\begin{table}[!htb]
\setlength{\tabcolsep}{4pt} 
\centering
\rowcolors{7}{rowblue}{white}
\resizebox{1.0\columnwidth}{!}{
\sisetup{detect-weight=true}
\begin{tabular}{lcccccccccccc}
\toprule
&  \multicolumn{4}{c}{3 views} &  \multicolumn{3}{c}{6 views} \\
\cmidrule(lr{0.5em}){2-5} \cmidrule(lr{0.5em}){6-8}
& 3DFAW & \multicolumn{2}{c}{H3DS} &  & \multicolumn{2}{c}{H3DS} &   \\
\cmidrule(lr{0.5em}){2-2} \cmidrule(lr{0.5em}){3-4} \cmidrule(lr{0.5em}){6-7}
& face $\downarrow$ & face $\downarrow$ & head $\downarrow$ & time $\downarrow$ & face $\downarrow$ & head $\downarrow$ & time $\downarrow$   \\
\cmidrule{1-8}
    MVF-Net                     & 1.61  & 1.67  & -     & \textbf{$\sim$ 1s}     & -     & -     & -     \\
    DFNRMVS                     & 1.74  & 1.96  & -     & \textbf{$\sim$ 1s}     & -     & -     & -     \\
    \cmidrule{1-8}
    \cmidrule{1-8}
    H3D-Net                     & \textbf{1.32}  & 1.47  & 11.13  & $\sim$ 20min     & 1.24  & \textbf{6.24}  & $\sim$ 40min     \\
    \textbf{Ours}               & \textbf{1.32}  & \textbf{1.39}  & \textbf{9.71}  & \textbf{$\sim$ 15s}  & \textbf{1.22}  & 8.03 & \textbf{$\sim$ 30s} \\
\bottomrule
\end{tabular}
}
\caption{{\bfseries 3D reconstruction method comparison (multi-view).} Error in millimeters. Time in seconds. Methods above the division line are 3DMM-based and methods below are model-free.}
\label{table:quantitative2}
\vspace{-1mm}
\end{table}

In Table \ref{table:quantitative2} we can see the multi-view results. We achieve better results compared with the parametric methods using 3 views, and for both view configurations we get comparable results to the implicit reconstruction methods, while being two orders of magnitude faster. Note that the normals estimation process takes less than 1 second, therefore it does not affect significantly the final time results.

\noindent{\bf Grid-based architecture efficiency.}
We propose a grid-based approach in order to speed up the query time of the network that models the SDF. In surface rendering, to find the intersection point it is needed to query multiple times along each ray. Therefore, replacing the monolithic MLP architecture by a grid-based approach with multiple trainable local features per grid and a shallow MLP, can significantly reduce this query time, as the local features indexing is significantly faster than computing a MLP forward with extra layers.

\vspace{-1mm}

\subsection{Qualitative results}

\vspace{-2mm}

\begin{figure}[]
    \centering
    \includegraphics[width=\columnwidth]{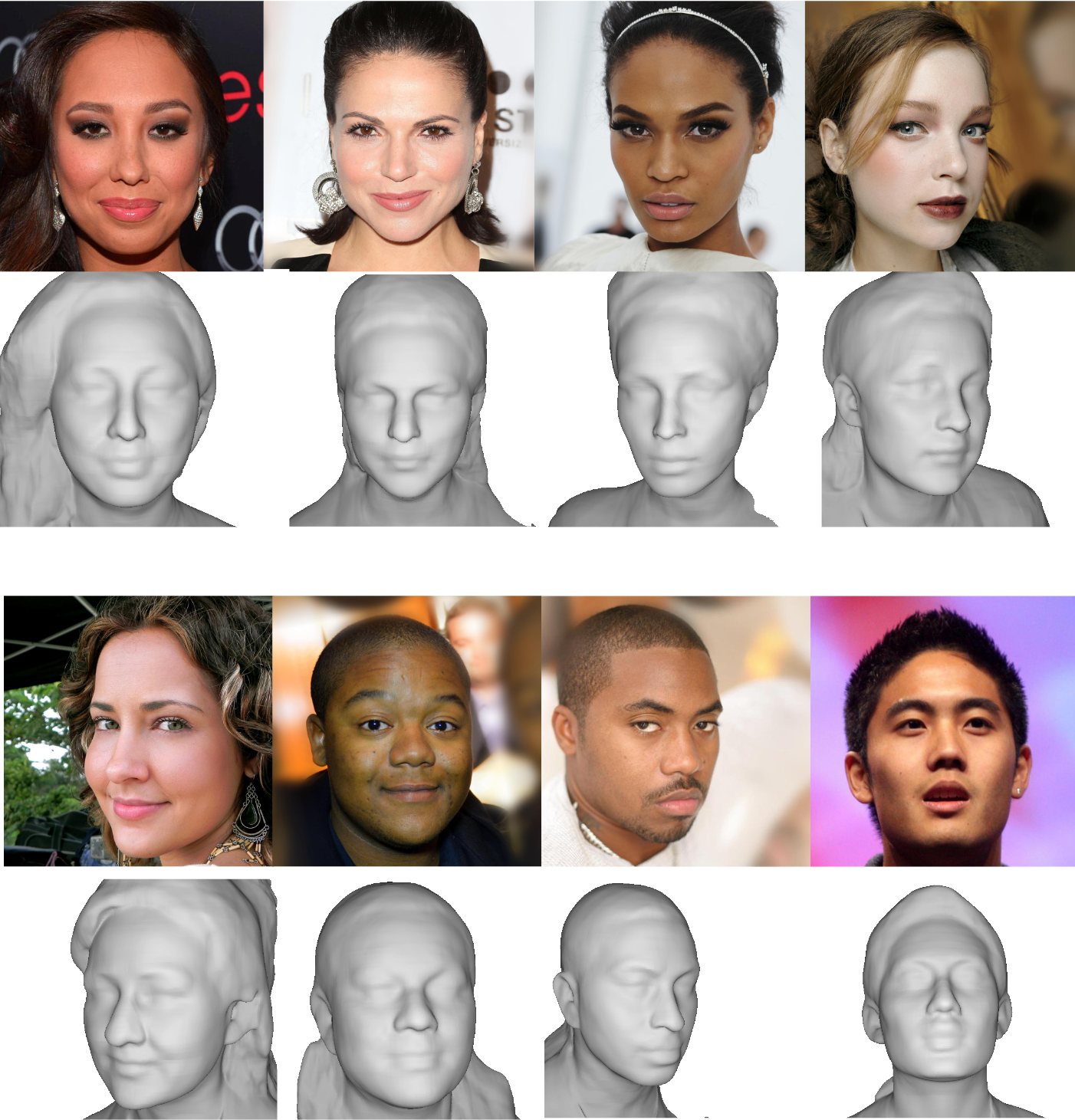}
    \caption{\textbf{Qualitative results.} \method{} results on CelebHQ dataset for a single input image.}
    \label{fig:qualitative_celeb}
    \vspace{-6mm}
\end{figure}

Qualitative results show comparable quality reconstruction in all the datasets while being up to 100x times faster. Our model is competitive in the one shot scenario, as well as in a few-shot setup, as shown in Figure \ref{fig:qualitative_experiments_front}. Unlike 3DMM-based approaches, which only provide an accurate reconstruction of the face region, \method{} is able to recover the identity by providing the geometry of the full head, including hair and shoulders.

\noindent
\textbf{Single-view reconstruction.} As expected, the SIRA model \cite{caselles2022sira} performs best in the single-image setting, as it is designed to achieve high accuracy in this configuration, but at the run-time cost of several minutes of reconstruction time. However, \method{} is able to obtain competitive visual results in a matter of seconds. On the other hand, the 3DMM-based models \cite{dib2021towards,feng2021learning} struggle to capture fine anatomical details and they don't model hair or shoulders as stated in \cite{Ramon_2021_ICCV}. Furthermore, we can see that our model is able to obtain more robust and natural results in areas such as the shoulders, which are challenging to estimate from a single view.

It is essential that the method is not biased and does not discriminate against any group, religion, colour, gender, age, or disability status. Therefore, we provide in Figure \ref{fig:qualitative_celeb} the results on the Celeb-HQ dataset composed of in-the-wild single images from the frontal view, to show that our model is diverse.

\noindent
\textbf{Multi-view reconstruction.} We evaluate the performance when increasing the number of input views on the reconstruction surface, in both the face and full head regions. As \method{} is a method that requires optimization per-scene, we consistently see improvement as the number of views increases. As we can see in Figure \ref{fig:qualitative_experiments_front}, we perform on par with the state-of-the-art implicit reconstruction method \cite{Ramon_2021_ICCV} in terms of Chamfer distance in both the face and full-head regions. Remarkably, \method{} is able to better reconstruct challenging high frequency regions such as the hair, as it can be seen in the 3th row of Figure \ref{fig:qualitative_experiments_front}.


\vspace{-3mm}

\section{Conclusions}
\label{sec:conclusions}

\vspace{-2mm}

We have demonstrated that it is possible to push the speed of implicit reconstruction methods based on neural fields by using a hybrid grid-based architecture in combination with surface rendering. We have shown that, while a naive combination of these two ideas leads to poor results due to unstable optimisation, these difficulties can be overcome by introducing appropriate geometry priors. In particular, we have demonstrated the effectiveness of pre-training the geometry network as a statistical shape model, and of using monocular normal predictions as cues, to stabilise the optimisation. Using these techniques, we achieve comparable state-of-the-art 3D reconstructions of human avatars with a 100x speedup over neural-field-based alternatives. This speedup can significantly ease its widespread adoption.

\vspace{-3mm}

\section{Limitations and future work}
\label{sec:future_work}

\vspace{-2mm}

First, an architecture based on dense grids may have some memory concerns, which have been discussed and handled already \cite{Takikawa_2021_CVPR} and was not at our primary scope, which was mainly speed. Also, it is also important to mention that the final representation capacity of our architecture is somehow limited by the quality of the normals predicted.


Second, despite its numerous advantages, grid representations combined with surface rendering still being an under-determined problem. In particular, in absence of a stronger inductive bias, we have showed how beneficial it is to guide this field with the help of prior knowledge. However, the commented prior architecture is limited in such a way that only a half of it - its decoder side - is optimized into its latent space for each different scene. The feature grid side of the prior, conversely, does not depend on the scene so it simply becomes a good initialization, which eases convergence, but does not let us make full use of it. Accordingly, an appropriate prior architecture which would let the grid local features to be optimized throughout its prior latent space, could potentially unlock the full capacity of these grid representations.

\vspace{-3mm}

\section{Acknowledgements}
\label{sec:acknowledgements}

\vspace{-2mm}

This work has been partially supported by: MoHuCo PID2020-120049RB-I00 funded by MCIN/AEI/10.13039/501100011033; DeeLight PID2020-117142GB-I00 funded by MCIN/ AEI /10.13039/501100011033; and the industrial doctorate 2021 DI 00102 funded by the Government of Catalonia.

\newpage
{\small
\bibliographystyle{ieee_fullname}
\bibliography{egbib}

\begin{thebibliography}{10}\itemsep=-1pt

\bibitem{Atzmon_2020_CVPR}
Matan Atzmon and Yaron Lipman.
\newblock Sal: Sign agnostic learning of shapes from raw data.
\newblock In {\em Proceedings of the IEEE/CVF Conference on Computer Vision and
  Pattern Recognition (CVPR)}, 2020.

\bibitem{bai2020deep}
Ziqian Bai, Zhaopeng Cui, Jamal~Ahmed Rahim, Xiaoming Liu, and Ping Tan.
\newblock Deep facial non-rigid multi-view stereo.
\newblock In {\em Proceedings of the IEEE/CVF Conference on Computer Vision and
  Pattern Recognition (CVPR)}, 2020.

\bibitem{besl1992method}
Paul~J Besl and Neil~D McKay.
\newblock A method for registration of 3-d shapes.
\newblock In {\em ACM Transactions on Graphics (TOG)}, 1992.

\bibitem{Brunton2014_Wavelets}
Alan Brunton, Timo Bolkart, and Stefanie Wuhrer.
\newblock Multilinear wavelets: A statistical shape space for human faces.
\newblock In {\em Proceedings of the IEEE/CVF European Conference on Computer
  Vision (ECCV)}, 2014.

\bibitem{burkov2022multi}
Egor Burkov, Ruslan Rakhimov, Aleksandr Safin, Evgeny Burnaev, and Victor
  Lempitsky.
\newblock Multi-neus: 3d head portraits from single image with neural implicit
  functions.
\newblock {\em arXiv preprint arXiv:2209.04436}, 2022.

\bibitem{cao2022authentic}
Chen Cao and Tomas Simon.
\newblock Authentic volumetric avatars from a phone scan.
\newblock {\em ACM Transactions on Graphics (TOG)}, 2022.

\bibitem{caselles2022sira}
Pol Caselles, Eduard Ramon, Jaime Garcia, Xavier Giro-i Nieto, Francesc
  Moreno-Noguer, and Gil Triginer.
\newblock Sira: Relightable avatars from a single image.
\newblock In {\em IEEE/CVF Winter Conference on Applications of Computer Vision
  (WACV)}, 2022.

\bibitem{REALY}
Zenghao Chai, Haoxian Zhang, Jing Ren, Di Kang, Zhengzhuo Xu, Xuefei Zhe, Chun
  Yuan, and Linchao Bao.
\newblock Realy: Rethinking the evaluation of 3d face reconstruction.
\newblock In {\em Proceedings of the IEEE/CVF European Conference on Computer
  Vision (ECCV)}, 2022.

\bibitem{Chatziagapi_3dv2021}
Aggelina Chatziagapi, ShahRukh Athar, Francesc Moreno-Noguer, and Dimitris
  Samaras.
\newblock {SIDER}: Single-image neural optimization for facial geometric detail
  recovery.
\newblock In {\em International Conference on 3D Vision (3DV)}, 2021.

\bibitem{dib2021towards}
Abdallah Dib, Cedric Thebault, Junghyun Ahn, Philippe-Henri Gosselin, Christian
  Theobalt, and Louis Chevallier.
\newblock Towards high fidelity monocular face reconstruction with rich
  reflectance using self-supervised learning and ray tracing.
\newblock In {\em Proceedings of the IEEE/CVF International Conference on
  Computer Vision (ICCV)}, 2021.

\bibitem{dou2018multi}
Pengfei Dou and Ioannis~A Kakadiaris.
\newblock Multi-view 3d face reconstruction with deep recurrent neural
  networks.
\newblock {\em Image and Vision Computing}, 80:80--91, 2018.

\bibitem{feng2021learning}
Yao Feng, Haiwen Feng, Michael~J Black, and Timo Bolkart.
\newblock Learning an animatable detailed 3d face model from in-the-wild
  images.
\newblock {\em ACM Transactions on Graphics (TOG)}, 40(4):1--13, 2021.

\bibitem{yu2022plenoxels}
Sara Fridovich-Keil, Alex Yu, Matthew Tancik, Qinhong Chen, Benjamin Recht, and
  Angjoo Kanazawa.
\newblock Plenoxels: Radiance fields without neural networks.
\newblock In {\em CVPR}, 2022.

\bibitem{grassal2022neural}
Philip-William Grassal, Malte Prinzler, Titus Leistner, Carsten Rother,
  Matthias Nie{\ss}ner, and Justus Thies.
\newblock Neural head avatars from monocular rgb videos.
\newblock In {\em Proceedings of the IEEE/CVF Conference on Computer Vision and
  Pattern Recognition (CVPR)}, 2022.

\bibitem{igr}
Amos Gropp, Lior Yariv, Niv Haim, Matan Atzmon, and Yaron Lipman.
\newblock Implicit geometric regularization for learning shapes.
\newblock In {\em Proceedings of Machine Learning and Systems (MLSys)}. 2020.

\bibitem{gropp2020implicit}
Amos Gropp, Lior Yariv, Niv Haim, Matan Atzmon, and Yaron Lipman.
\newblock Implicit geometric regularization for learning shapes.
\newblock {\em arXiv preprint arXiv:2002.10099}, 2020.

\bibitem{karras2017progressive}
Tero Karras, Timo Aila, Samuli Laine, and Jaakko Lehtinen.
\newblock Progressive growing of gans for improved quality, stability, and
  variation.
\newblock {\em arXiv preprint arXiv:1710.10196}, 2017.

\bibitem{kellnhofer2021neural}
Petr Kellnhofer, Lars~C Jebe, Andrew Jones, Ryan Spicer, Kari Pulli, and Gordon
  Wetzstein.
\newblock Neural lumigraph rendering.
\newblock In {\em Proceedings of the IEEE/CVF Conference on Computer Vision and
  Pattern Recognition (CVPR)}, 2021.

\bibitem{kingma2014adam}
Diederik~P Kingma and Jimmy Ba.
\newblock Adam: A method for stochastic optimization.
\newblock {\em arXiv preprint arXiv:1412.6980}, 2014.

\bibitem{li2022vox}
Hai Li, Xingrui Yang, Hongjia Zhai, Yuqian Liu, Hujun Bao, and Guofeng Zhang.
\newblock Vox-surf: Voxel-based implicit surface representation.
\newblock {\em arXiv preprint arXiv:2208.10925}, 2022.

\bibitem{li2023nerfacc}
Ruilong Li, Hang Gao, Matthew Tancik, and Angjoo Kanazawa.
\newblock Nerfacc: Efficient sampling accelerates nerfs.
\newblock {\em arXiv preprint arXiv:2305.04966}, 2023.

\bibitem{li2017learning}
Tianye Li, Timo Bolkart, Michael~J Black, Hao Li, and Javier Romero.
\newblock Learning a model of facial shape and expression from 4d scans.
\newblock {\em ACM Transactions on Graphics (TOG)}, 36(6):194--1, 2017.

\bibitem{FLAME:SiggraphAsia2017}
Tianye Li, Timo Bolkart, Michael.~J. Black, Hao Li, and Javier Romero.
\newblock Learning a model of facial shape and expression from {4D} scans.
\newblock {\em ACM Transactions on Graphics (TOG)}, 36(6):194:1--194:17, 2017.

\bibitem{lin2021barf}
Chen-Hsuan Lin, Wei-Chiu Ma, Antonio Torralba, and Simon Lucey.
\newblock Barf: Bundle-adjusting neural radiance fields.
\newblock In {\em Proceedings of the IEEE/CVF International Conference on
  Computer Vision (ICCV)}, 2021.

\bibitem{Lin_2020_CVPR}
Jiangke Lin, Yi Yuan, Tianjia Shao, and Kun Zhou.
\newblock Towards high-fidelity 3d face reconstruction from in-the-wild images
  using graph convolutional networks.
\newblock In {\em Proceedings of the IEEE/CVF Conference on Computer Vision and
  Pattern Recognition (CVPR)}, 2020.

\bibitem{Lindell20arxiv_AutoInt}
David Lindell, Julien Martel, and Gordon Wetzstein.
\newblock {AutoInt}: Automatic integration for fast neural volume rendering.
\newblock {\em arXiv preprint arXiv:2012.01714}, 2020.

\bibitem{lindell2022bacon}
David~B Lindell, Dave Van~Veen, Jeong~Joon Park, and Gordon Wetzstein.
\newblock Bacon: Band-limited coordinate networks for multiscale scene
  representation.
\newblock In {\em Proceedings of the IEEE/CVF Conference on Computer Vision and
  Pattern Recognition (CVPR)}, 2022.

\bibitem{littwin2019deep}
Gidi Littwin and Lior Wolf.
\newblock Deep meta functionals for shape representation.
\newblock In {\em Proceedings of the IEEE/CVF International Conference on
  Computer Vision (ICCV)}, 2019.

\bibitem{SMPL:2015}
Matthew Loper, Naureen Mahmood, Javier Romero, Gerard Pons-Moll, and Michael~J.
  Black.
\newblock {SMPL}: A skinned multi-person linear model.
\newblock {\em ACM Transactions on Graphics (TOG)}, 34(6):248:1--248:16, 2015.

\bibitem{mescheder2019occupancy}
Lars Mescheder, Michael Oechsle, Michael Niemeyer, Sebastian Nowozin, and
  Andreas Geiger.
\newblock Occupancy networks: Learning 3d reconstruction in function space.
\newblock In {\em Proceedings of the IEEE/CVF Conference on Computer Vision and
  Pattern Recognition (CVPR)}, 2019.

\bibitem{mildenhall2020nerf}
Ben Mildenhall, Pratul~P. Srinivasan, Matthew Tancik, Jonathan~T. Barron, Ravi
  Ramamoorthi, and Ren Ng.
\newblock Nerf: Representing scenes as neural radiance fields for view
  synthesis.
\newblock In {\em Proceedings of the IEEE/CVF European Conference on Computer
  Vision (ECCV)}, 2020.

\bibitem{muller2022instant}
Thomas M{\"u}ller, Alex Evans, Christoph Schied, and Alexander Keller.
\newblock Instant neural graphics primitives with a multiresolution hash
  encoding.
\newblock {\em arXiv preprint arXiv:2201.05989}, 2022.

\bibitem{mueller2022instant}
Thomas M\"uller, Alex Evans, Christoph Schied, and Alexander Keller.
\newblock Instant neural graphics primitives with a multiresolution hash
  encoding.
\newblock {\em ACM Transactions on Graphics (TOG)}, 41(4):102:1--102:15, 2022.

\bibitem{niemeyer2020differentiable}
Michael Niemeyer, Lars Mescheder, Michael Oechsle, and Andreas Geiger.
\newblock Differentiable volumetric rendering: Learning implicit 3d
  representations without 3d supervision.
\newblock In {\em Proceedings of the IEEE/CVF Conference on Computer Vision and
  Pattern Recognition (CVPR)}, 2020.

\bibitem{oechsle2021unisurf}
Michael Oechsle, Songyou Peng, and Andreas Geiger.
\newblock Unisurf: Unifying neural implicit surfaces and radiance fields for
  multi-view reconstruction.
\newblock In {\em Proceedings of the IEEE/CVF International Conference on
  Computer Vision (ICCV)}, 2021.

\bibitem{park2019deepsdf}
Jeong~Joon Park, Peter Florence, Julian Straub, Richard Newcombe, and Steven
  Lovegrove.
\newblock Deepsdf: Learning continuous signed distance functions for shape
  representation.
\newblock In {\em Proceedings of the IEEE/CVF Conference on Computer Vision and
  Pattern Recognition (CVPR)}, 2019.

\bibitem{bfm09}
P. Paysan, R. Knothe, B. Amberg, S. Romdhani, and T. Vetter.
\newblock A 3d face model for pose and illumination invariant face recognition.
\newblock In {\em Proceedings of the 6th IEEE International Conference on
  Advanced Video and Signal based Surveillance (AVSS) for Security, Safety and
  Monitoring in Smart Environments}, 2009.

\bibitem{pillai20192nd}
Rohith~Krishnan Pillai, L{\'a}szl{\'o}~Attila Jeni, Huiyuan Yang, Zheng Zhang,
  Lijun Yin, and Jeffrey~F Cohn.
\newblock The 2nd 3d face alignment in the wild challenge (3dfaw-video): Dense
  reconstruction from video.
\newblock In {\em Proceedings of the IEEE/CVF International Conference on
  Computer Vision Workshops (ICCV Workshops)}, 2019.

\bibitem{pumarolavisco}
Albert Pumarola, Artsiom Sanakoyeu, Lior Yariv, Ali Thabet, and Yaron Lipman.
\newblock Visco grids: Surface reconstruction with viscosity and coarea grids.
\newblock In {\em Advances in Neural Information Processing Systems (NeurIPS)}.

\bibitem{ramon2019multi}
Eduard Ramon, Janna Escur, and Xavier Giro-i Nieto.
\newblock Multi-view 3d face reconstruction in the wild using siamese networks.
\newblock In {\em Proceedings of the IEEE/CVF International Conference on
  Computer Vision Workshops (ICCV Workshops)}, 2019.

\bibitem{Ramon_2021_ICCV}
Eduard Ramon, Gil Triginer, Janna Escur, Albert Pumarola, Jaime Garcia, Xavier
  Gir\'o-i Nieto, and Francesc Moreno-Noguer.
\newblock H3d-net: Few-shot high-fidelity 3d head reconstruction.
\newblock In {\em Proceedings of the IEEE/CVF International Conference on
  Computer Vision (ICCV)}, 2021.

\bibitem{richardson20163d}
Elad Richardson, Matan Sela, and Ron Kimmel.
\newblock 3d face reconstruction by learning from synthetic data.
\newblock In {\em International Conference on 3D Vision (3DV)}, 2016.

\bibitem{richardson2017learning}
Elad Richardson, Matan Sela, Roy Or-El, and Ron Kimmel.
\newblock Learning detailed face reconstruction from a single image.
\newblock In {\em Proceedings of the IEEE/CVF Conference on Computer Vision and
  Pattern Recognition (CVPR)}, 2017.

\bibitem{sitzmann2019metasdf}
Vincent Sitzmann, Eric~R. Chan, Richard Tucker, Noah Snavely, and Gordon
  Wetzstein.
\newblock Metasdf: Meta-learning signed distance functions.
\newblock In {\em Advances in Neural Information Processing Systems (NeurIPS)},
  2020.

\bibitem{smith2020morphable}
William A.~P. Smith, Alassane Seck, Hannah Dee, Bernard Tiddeman, Joshua
  Tenenbaum, and Bernhard Egger.
\newblock A morphable face albedo model.
\newblock In {\em Proceedings of the IEEE/CVF Conference on Computer Vision and
  Pattern Recognition (CVPR)}, 2020.

\bibitem{sun2022direct}
Cheng Sun, Min Sun, and Hwann-Tzong Chen.
\newblock Direct voxel grid optimization: Super-fast convergence for radiance
  fields reconstruction.
\newblock In {\em Proceedings of the IEEE/CVF Conference on Computer Vision and
  Pattern Recognition (CVPR)}, 2022.

\bibitem{Takikawa_2021_CVPR}
Towaki Takikawa, Joey Litalien, Kangxue Yin, Karsten Kreis, Charles Loop, Derek
  Nowrouzezahrai, Alec Jacobson, Morgan McGuire, and Sanja Fidler.
\newblock Neural geometric level of detail: Real-time rendering with implicit
  3d shapes.
\newblock In {\em Proceedings of the IEEE/CVF Conference on Computer Vision and
  Pattern Recognition (CVPR)}, 2021.

\bibitem{pmlr-v97-tan19a}
Mingxing Tan and Quoc Le.
\newblock Efficientnet: Rethinking model scaling for convolutional neural
  networks.
\newblock In {\em Proceedings of the 36th International Conference on Machine
  Learning (ICML)}, 2019.

\bibitem{tancik2020meta}
Matthew Tancik, Ben Mildenhall, Terrance Wang, Divi Schmidt, Pratul~P.
  Srinivasan, Jonathan~T. Barron, and Ren Ng.
\newblock Learned initializations for optimizing coordinate-based neural
  representations.
\newblock In {\em Proceedings of the IEEE/CVF Conference on Computer Vision and
  Pattern Recognition (CVPR)}, 2021.

\bibitem{tancik2020fourfeat}
Matthew Tancik, Pratul~P. Srinivasan, Ben Mildenhall, Sara Fridovich-Keil,
  Nithin Raghavan, Utkarsh Singhal, Ravi Ramamoorthi, Jonathan~T. Barron, and
  Ren Ng.
\newblock Fourier features let networks learn high frequency functions in low
  dimensional domains.
\newblock {\em Advances in Neural Information Processing Systems (NeurIPS)},
  2020.

\bibitem{tewari2017mofa}
Ayush Tewari, Michael Zollhofer, Hyeongwoo Kim, Pablo Garrido, Florian Bernard,
  Patrick Perez, and Christian Theobalt.
\newblock Mofa: Model-based deep convolutional face autoencoder for
  unsupervised monocular reconstruction.
\newblock In {\em Proceedings of the IEEE/CVF International Conference on
  Computer Vision (ICCV)}, 2017.

\bibitem{tran2018extreme}
Anh~Tuan Tran, Tal Hassner, Iacopo Masi, Eran Paz, Yuval Nirkin, and
  G{\'e}rard~G Medioni.
\newblock Extreme 3d face reconstruction: Seeing through occlusions.
\newblock In {\em Proceedings of the IEEE/CVF Conference on Computer Vision and
  Pattern Recognition (CVPR)}, 2018.

\bibitem{tuan2017regressing}
Anh Tuan~Tran, Tal Hassner, Iacopo Masi, and G{\'e}rard Medioni.
\newblock Regressing robust and discriminative 3d morphable models with a very
  deep neural network.
\newblock In {\em Proceedings of the IEEE/CVF Conference on Computer Vision and
  Pattern Recognition (CVPR)}, 2017.

\bibitem{wang2022faceverse}
Lizhen Wang, Zhiyua Chen, Tao Yu, Chenguang Ma, Liang Li, and Yebin Liu.
\newblock Faceverse: a fine-grained and detail-controllable 3d face morphable
  model from a hybrid dataset.
\newblock In {\em Proceedings of the IEEE/CVF Conference on Computer Vision and
  Pattern Recognition (CVPR)}, 2022.

\bibitem{wang2021neus}
Peng Wang, Lingjie Liu, Yuan Liu, Christian Theobalt, Taku Komura, and Wenping
  Wang.
\newblock Neus: Learning neural implicit surfaces by volume rendering for
  multi-view reconstruction.
\newblock {\em arXiv preprint arXiv:2106.10689}, 2021.

\bibitem{neus2}
Yiming Wang, Qin Han, Marc Habermann, Kostas Daniilidis, Christian Theobalt,
  and Lingjie Liu.
\newblock Neus2: Fast learning of neural implicit surfaces for multi-view
  reconstruction.
\newblock In {\em Proceedings of the IEEE/CVF International Conference on
  Computer Vision (ICCV)}, 2023.

\bibitem{wu2019mvf}
Fanzi Wu, Linchao Bao, Yajing Chen, Yonggen Ling, Yibing Song, Songnan Li,
  King~Ngi Ngan, and Wei Liu.
\newblock Mvf-net: Multi-view 3d face morphable model regression.
\newblock In {\em Proceedings of the IEEE/CVF Conference on Computer Vision and
  Pattern Recognition (CVPR)}, 2019.

\bibitem{yariv2021volume}
Lior Yariv, Jiatao Gu, Yoni Kasten, and Yaron Lipman.
\newblock Volume rendering of neural implicit surfaces.
\newblock {\em Advances in Neural Information Processing Systems (NeurIPS)},
  2021.

\bibitem{yariv2020multiview}
Lior Yariv, Yoni Kasten, Dror Moran, Meirav Galun, Matan Atzmon, Basri Ronen,
  and Yaron Lipman.
\newblock Multiview neural surface reconstruction by disentangling geometry and
  appearance.
\newblock {\em Advances in Neural Information Processing Systems (NeurIPS)},
  2020.

\bibitem{yu2021plenoctrees}
Alex Yu, Ruilong Li, Matthew Tancik, Hao Li, Ren Ng, and Angjoo Kanazawa.
\newblock {PlenOctrees} for real-time rendering of neural radiance fields.
\newblock In {\em ICCV}, 2021.

\bibitem{Yu2022MonoSDF}
Zehao Yu, Songyou Peng, Michael Niemeyer, Torsten Sattler, and Andreas Geiger.
\newblock Monosdf: Exploring monocular geometric cues for neural implicit
  surface reconstruction.
\newblock {\em Advances in Neural Information Processing Systems (NeurIPS)},
  2022.

\bibitem{zheng2022avatar}
Yufeng Zheng, Victoria~Fern{\'a}ndez Abrevaya, Marcel~C B{\"u}hler, Xu Chen,
  Michael~J Black, and Otmar Hilliges.
\newblock Im avatar: Implicit morphable head avatars from videos.
\newblock In {\em Proceedings of the IEEE/CVF Conference on Computer Vision and
  Pattern Recognition (CVPR)}, 2022.

\bibitem{zhou2018unetplusplus}
Zongwei Zhou, Md~Mahfuzur~Rahman Siddiquee, Nima Tajbakhsh, and Jianming Liang.
\newblock Unet++: A nested u-net architecture for medical image segmentation.
\newblock In {\em Deep Learning in Medical Image Analysis and Multimodal
  Learning for Clinical Decision Support}. Springer, 2018.

\end{thebibliography}
}


\end{document}